\documentclass{midl} % Include author names
\usepackage{mwe} % to get dummy images
\usepackage{algorithm2e}
\usepackage{color}
\usepackage{multirow}
\jmlrvolume{-- Under Review}
\jmlryear{2019}
\jmlrworkshop{Full Paper -- MIDL 2019 accepted}
\title[Superpixel-based Data Augmentation]{SPDA: Superpixel-based Data Augmentation for\\Biomedical Image Segmentation}

\midlauthor{\Name{Yizhe Zhang\nametag{$^{1}$}} \Email{yzhang29@nd.edu}\\
\Name{Lin Yang\nametag{$^{1}$}} \Email{lyang5@nd.edu}\\
\Name{Hao Zheng\nametag{$^{1}$}} \Email{hzheng3@nd.edu}\\
\Name{Peixian Liang\nametag{$^{1}$}} \Email{pliang@nd.edu}\\
\Name{Colleen Mangold\nametag{$^{2}$}} \Email{cav154@psu.edu}\\
\Name{Raquel G. Loreto\nametag{$^{2}$}} \Email{raquelgloreto@gmail.com}\\
\Name{David P. Hughes\nametag{$^{2}$}} \Email{dhughes@psu.edu}\\
\Name{Danny Z. Chen\nametag{$^{1}$}} \Email{dchen@nd.edu}\\
\addr $^{1}$Department of Computer Science and Engineering, University of Notre Dame, USA \\
\addr $^{2}$Department of Entomology and Department of Biology, Center for Infectious Disease Dynamics, Pennsylvania State University, USA\\
}

\begin{document}

\maketitle

\begin{abstract}

{\color{black}{Supervised training a deep neural network aims to ``teach" the network to mimic human visual perception that is represented by image-and-label pairs in the training data. Superpixelized (SP) images are visually perceivable to humans, but a conventionally trained deep learning model often performs poorly when working on SP images. To better mimic human visual perception, we think it is desirable for the deep learning model to be able to perceive not only raw images but also SP images. In this paper, we propose a new superpixel-based data augmentation (SPDA) method for training deep learning models for biomedical image segmentation. Our method applies a superpixel generation scheme to all the original training images to generate superpixelized images. The SP images thus obtained are then jointly used with the original training images to train a deep learning model. Our experiments of SPDA on four biomedical image datasets show that SPDA is effective and can consistently improve the performance of state-of-the-art fully convolutional networks for biomedical image segmentation in 2D and 3D images. Additional studies also demonstrate that SPDA can practically reduce the generalization gap.}}

%Superpixels are over-segmented image regions that are perceptually meaningful and approximately complete. Superpixels can reduce the complexity of images and represent images in an effective and compact way. Information such as shapes, colors, and object-level relations is well captured and represented by superpixels. 

%in this paper, we propose a new data augmentation method using superpixels for training deep learning segmentation models.

\end{abstract}

\section{Introduction}

%Data augmentation is widely used to artificially inflate the training datasets. Due to the fact that labeled training data, especially in biomedical imaging \cite{weese2016four}, are usually expensive and difficult to acquire, data augmentation is highly useful and necessary for training deep learning model for biomedical image segmentation tasks.

Traditional data augmentation methods use a combination of geometric transformations to artificially inflate training data \cite{perez2017effectiveness}. For each raw training image and its corresponding annotated image, it generates ``duplicate'' images that are shifted, zoomed in/out, rotated, flipped, and/or distorted. These basic/traditional data augmentation methods are generally applicable to classification problems where the output is a vector and segmentation problems where the output is a segmentation map. %When applying to a classification problem, basic geometric transform on an raw image usually do not affect its class label, thus, the class label remains the same when applying geometric transform based data augmentation. When these transformations are applied to the segmentation problem, a transform (e.g. rotation) applied on an raw image should be also applied on its corresponding label map. This ensures that the relation between the raw image and its label map is accurately maintained during data augmentation.

%A natural question arises as whether image domain cues (e.g., local pixel-wise similarity/dissimilarity) can be explored for data augmentation. 
{\color{black}{Recently, generative adversarial networks (GANs) have been used for data augmentation (e.g., \cite{antoniou2017data}). Encouraging the generator to produce realistic looking images (comparing to the original images) is a main consideration when training the generator. A key issue to this consideration is that it does not define/imply what kind of generated images would be useful/meaningful for data augmentation purpose, and the generator does not necessarily converge to a model version that generates useful new data for training a better segmentation or classification model. \cite{wang2018low} was proposed to deal with this issue using a task-related classifier for training an image generator. However, the method in \cite{wang2018low} was designed for classification problems; in segmentation, the distributions of labels are usually much more complicated and it is quite non-trivial to extend the method \cite{wang2018low} to segmentation tasks.}} % A recent study \cite{perez2017effectiveness} also shows GAN performs worse than traditional data augmentation method in some cases. It is possible that they may not tune the GAN model well enough, but this at least suggests that GAN-type DA is not very easy to tune/use.

\begin{figure*}[t]
  \centering
  \includegraphics[width=4.4 in]{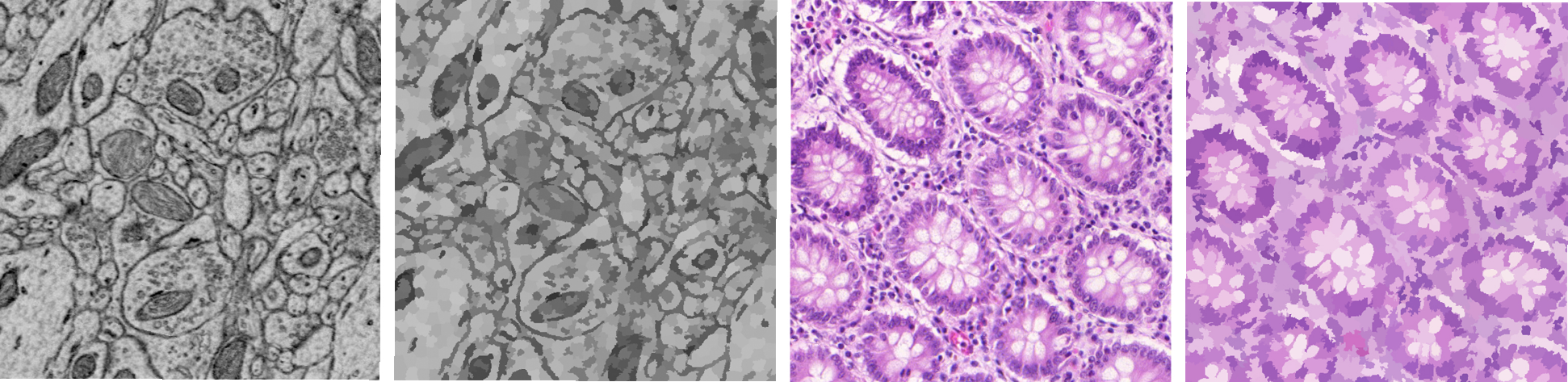}
  \caption{From left to right: An electron micrograph of neuronal structure, its superpixelized image, an H\&E stained pathological image of glands, and its superpixelized image. The superpixels preserve the essential objects and their boundaries.}
\label{fig:sp_demo}
\end{figure*}

{\color{black}{As an algorithm based (non-learning based) data augmentation technique, mixup \cite{zhang2017mixup} was proposed to generate new image samples ``between'' pairs of training samples for image classification problems. It was motivated based on the principles of Vicinal Risk Minimization \cite{chapelle2001vicinal} and its experimental results showed promising classification accuracy improvement. In \cite{eaton2018improving}, it extended the mixup method to medical image segmentation, showing that mixup is also applicable to data augmentation for segmentation problems. 

In this paper, we propose a new algorithm-based data augmentation technique that uses superpixels for better training a deep learning model for biomedical image segmentation. Our method is based on a common experience that superpixelized (SP) images are visually perceivable to humans (see Fig.~\ref{fig:sp_demo}), but a conventionally trained deep learning model (trained using only raw images) often performs poorly when working on SP images. This phenomenon implies that a conventionally trained deep learning model may not mimic human visual behaviors well enough. Thus, we think encouraging a deep learning network to be able to perceive not only raw images but also SP images can make it more closely mimic human visual perception. Our method is built on this idea, by adding SP images to the training data for training a deep learning model. Our new superpixel-based data augmentation (SPDA) method can work together with traditional data augmentation methods and be generally applicable to many deep learning based image segmentation models.}}

A short summary of our SPDA method is as follows. For each raw image, we apply a superpixel generation method (e.g., SLIC \cite{achanta2012slic}) to obtain superpixel cells. Superpixel cells are groups of pixels that are visually similar and spatially connected. For every superpixel cell $C$, we compute the average pixel value(s) for all the pixels in $C$ and assign the computed average value(s) to all the pixels in $C$. In this way, we effectively remove very local image details and emphasize more on the overall colors, shapes, and spatial relations of objects in the image (see Fig.~\ref{fig:sp_demo}). After ``superpixelizing'' all the raw images in the original training set, we put all the superpixelized (SP) images into the training set together with the original training images for training a deep learning model. Our experiments of SPDA on four biomedical image datasets show that SPDA is effective and can consistently improve the performance of state-of-the-art fully convolutional networks (FCNs) for biomedical image segmentation in 2D and 3D images.

In Section \ref{sec:2}, we discuss several technical considerations on generating superpixelized images for data augmentation, and present our exact procedure for generating and using superpixelized images to train deep learning models. In Section \ref{sec:3}, we evaluate SPDA using multiple widely used FCNs on four biomedical image segmentation datasets, and show that SPDA consistently yields segmentation performance improvement on these datasets. 

%From the deep learning perspective, SPDA helps to reduce the complexity of the original data space to a simpler space and training with data from the simpler space essentially serves a regularization and xxx. 

%Superpixels are over-segmented image regions which are perceptually meaningful and near-
%complete. Superpixels can reduce the complexity of images and can represent the original image in an effective and compact way. Information such as shapes and object-level relations are well captured and represented using superpixels. SLIC is the most widely used superpixel generation methods and recently, xxx xxx xxx are proposed for generating better quality superpixels (higher boundary recall with less number of superpixels).

%Shifting, zooming in/out, rotating and flipping are often safe to use in biomedical image segmentation. As nowadays the state-of-the-art models, e.g. CNN, FCN, are window based, shifting/cropping process is widely used when train and apply an FCN model. Zooming in/out is useful to model the image variations caused by scales. 

%The performance xxx and is not very xxx.(perform not as good as the traditional approach) In this paper, we aim to make use of superpixels as a xxx and xxx data transformations for manipulating the training data for data augmentation.

\section{Superpixels for Data Augmentation}\label{sec:2}
\begin{figure*}[]
\centering
  \includegraphics[width= 5.0 in]{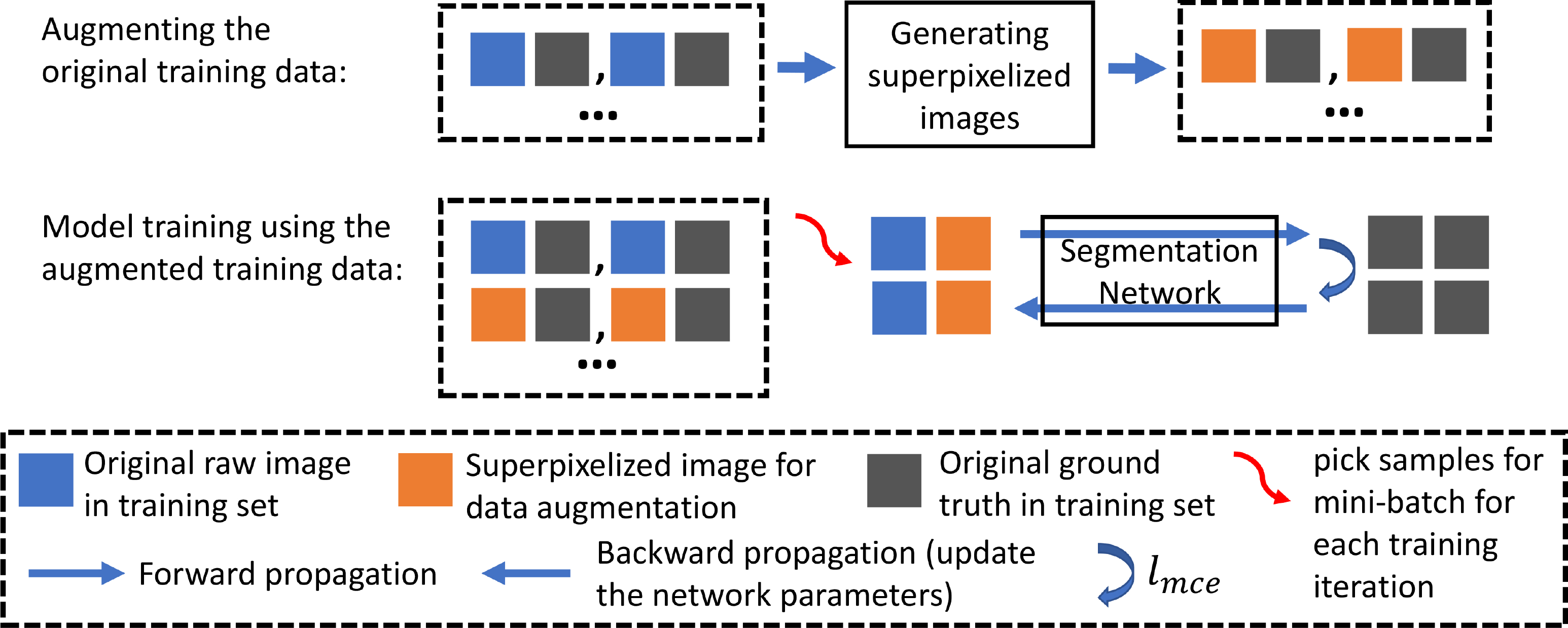}
  \caption{An overview of our SPDA method. During training, we first generate superpixelized images for all the raw images, and then add the SPDA-generated data to the training data for training a segmentation model. Note that the trained network is applied to only raw images during model testing.}
\label{fig:method_overview}
\end{figure*}

First, we give some notation and background of data augmentation. Then, we discuss several technical considerations on using superpixels for data augmentation. Finally, we present the key technical components: (i) What superpixel generation method we choose to use and the logic behind it; (ii) the exact procedure for generating superpixelized images; (iii) the training objective function and algorithm for using SPDA-generated images in deep learning model training. Fig.~\ref{fig:method_overview} gives an overview of our SPDA method for model training.

\subsection{Notation and preliminaries}
Given a set of image samples $X=\{x_1, \ldots,x_n\}$ and their corresponding ground truth $Y=\{y_1, \ldots, y_n\}$, for training a segmentation model (e.g., an FCN) $f \in F$ that describes the relationship between $x_i$ and $y_i$, the empirical risk is: 
\begin{equation}\label{eq1}
\frac{1}{n}\sum_{i=1}^{n}\mathcal{L}(f(x_i),y_i)
\end{equation}

\noindent
where $\mathcal{L}$ is a loss function (e.g., the cross-entropy). Learning the function $f$ is by minimizing Eq.~(\ref{eq1}), which is also known as Empirical Risk Minimization.  

One could use some proper functions to generate more data based on the original training data pair $(x_i, y_i)$. In general, we denote the generated data by $(x_i^{aug}, y_i^{aug})$.
%When newly generated data are used together with the original data, a simple form of %the new objective function may be:
%\begin{equation}\label{eq2}
%\frac{1}{n}\sum_{i=1}^{n}(\mathcal{L}(f(x_i),y_i) + \lambda %\mathcal{L}(f(x_i^{aug}),y_i^{aug}))
%\end{equation}
%\noindent
When there are multiple ($k$) versions of augmented data for one pair $(x_i,y_i)$, the loss with augmented data can be written as:
\begin{equation}\label{eq3}
\frac{1}{n}\sum_{i=1}^{n}(\mathcal{L}(f(x_i),y_i) + \lambda \sum_{j=1}^{k}\mathcal{L}(f(x_i^{aug_j}),y_i^{aug_j}))
\end{equation}

\noindent
where $\lambda$ is a hyper-parameter that controls the importance of the data augmentation term. Different ways of data augmentation produce different new data, and thus directly affect the learning procedure of $f$. As a common practice, flipping, rotation, cropping, etc. are widely used for data augmentation. This type of data augmentation applies geometric transformations ($g_k$, for $k$ different geometric transformations) to both $x_i$ and $y_i$, to generate new pairs of training data. For this type of data augmentation, Eq.~(\ref{eq3}) can be rewritten as: 
\begin{equation}\label{eq4} 
\frac{1}{n}\sum_{i=1}^{n}(\mathcal{L}(f(x_i),y_i) + \lambda \sum_{j=1}^{k}\mathcal{L}(f(g_j(x_i)),g_j(y_i)))
\end{equation}

Another type of data augmentation makes no change on $y_i$, and the only modification/augmentation is on $x_i$ (e.g., color jittering \cite{krizhevsky2012imagenet}). For this type of augmentation, Eq.~(\ref{eq3}) can simply be:
\begin{equation}\label{eq5} 
\frac{1}{n}\sum_{i=1}^{n}(\mathcal{L}(f(x_i),y_i) + \lambda \sum_{j=1}^{k}\mathcal{L}(f(G(x_i)),y_i))
\end{equation}
\noindent
where $G(\cdot)$ is a label-preserving transformation. Our new SPDA method belongs to this category. We propose to generate superpixlized images (denoted by $SP(\cdot)$) as a type of label-preserving (perception-preserving) transformation for data augmentation. 

Below we discuss several technical considerations on using superpixels for data augmentation, the technical details of $SP(\cdot)$, and how to use SPDA-generated data for model training.

%With augmented data added in the loss function, the feasible hypothesis space will be better shaped and regulated so that the learned model can generalized better on unseen images (overfitting problem can be xxx).

%The training (learning) procedure update the model parameters to minmize xxx. Statistically, the training procedure finds a ways to explain... It learns features that related to xxx and for noises and xxx not related to ..., one expect it to ...

%We aim to generate novel augmented data that can help to train a model which is, in certain level, invariant to local image changes (noises), meanwhile without any force of reducing the model's complexity. Adding such augmented data can help reduce the variance errors that in the bias–variance tradeoff without suffering higher bias errors.

%In order to train the model that is invariant to local pixel changes (noises), we first need to be clear that what image feature we expect the model be invariant of, and what image features we expect the model to pick up (related to the xxx). 

%make use of more higher image level cues. Such augmented data should improve the segmentation model's capability in handling possible low level pixel changes changes in unseen images. In figure xxx, we illustrated examples of possible xxx between training data and future unseen data. 

\subsection{Technical considerations}\label{why}
In this subsection, we discuss three technical considerations on generating superpixelized images for data augmentation.

%(1) for training a model that is robust/insensitive to small pixel value changes; (2) Adding superpixelized images encourages the model to also take care of the locations near the original data point.  (3) Superpixelized images' data distribution is perceptionally closer to the test data distribution than the original training data. Adding them to the training set can help reduce the well known generalization gap.

(1) Superpixelizing an image $x$ removes or reduces local image details in $x$ that might be less relevant to modeling $P(y|x,\theta_{f})$ ($\theta_{f}$ denotes the parameters of the segmentation model $f$). A superpixelized image $SP(x)$ is a simplified version of the original image $x$. Letting a deep learning model learn from $SP(x)$ to predict $y$ means asking the model to use little or no local (insignificant) pixel value changes and focus more on higher-level semantic information. Since model parameters are shared between predicting $y$ when given $x$ and predicting $y$ when given $SP(x)$, modeling $P(y|SP(x),\theta_{f})$ will influence modeling $P(y|x,\theta_{f})$. As a result, because of the joint modeling of $P(y|SP(x),\theta_{f})$, the learned function for predicting $y$ given $x$ would become more invariant/insensitive to local image noise and small details, and would learn and utilize more higher level image information and representations. \textcolor{black}{Note that all the original training images with all their local image details are still fully kept in the training dataset. Hence, whenever needed, the learning procedure is still able to use any local image details for modeling $P(y|x,\theta_{f})$. } % On the other hand, thanks to the added SPDA data, the learning procedure will also be able to learn and utilize more robust middle or higher level image representation.}

%(see Fig.~\ref{fig:sp_demo}). 
%Intuitively, superpixelized images are ``readable" to humans. Although 

%Technically, superpixels are over-segmented image regions that preserve essential object structures in an original image. Although ``superpixelizing'' an image removes local (insignificant) pixel changes in the image, a superpixelized image still contains essential information for predicting $y$.

%Adding xxx can be viewed as a regularization using augmented data.

(2) SPDA provides new image samples that are ``close" to the original training samples.
Under the principle of Vicinal Risk Minimization or VRM \cite{chapelle2001vicinal}, a vicinity or neighborhood around every training sample is defined or suggested  based on human knowledge. Additional samples then can be drawn from this vicinity distribution of the training samples to increase or enlarge the support of the training sample distribution \cite{zhang2017mixup}. Superpixelized images most of the time are conceptually meaningful to human eyes. It is likely that superpixelized images are also close\footnote{Being close means the distance (e.g., Euclidean distance) between an SPDA-generated image and its corresponding raw image is smaller than the distance between this raw image and any other raw image.} to their corresponding original image samples in the data space. If this ``close neighborhood" property is true, then adding SPDA-generated data to the training data should be helpful to improve the generalization capability of the model, according to VRM \cite{chapelle2001vicinal}. In {\bf Appendix \ref{EmpiricalStudy-1}}, we show that after using a generic dimensionality reduction method (e.g., PCA, t-SNE \cite{maaten2008visualizing}), one can observe that each superpixelized image is in a close neighborhood of its corresponding original image. 

% can confirm/show that superpixelized images are indeed in a close neighborhood of the original training samples, then it is almost certain that adding superpixelized images to the training set can be helpful for reducing the generalization gap, based on the principle of Vicinal Risk Minimization. 

(3) Adding superpixelized images to the training set makes the data distribution of the training set thus resulted closer to the test data distribution or the true data distribution.
Superpixelized images form a more general and broader base for the visual conception related to the learning task. Adding superpixelized images to the original training data makes the training data distribution have a more generic base that can potentially better support unseen test images. In {\bf Appendix \ref{EmpiricalStudy-2}}, using variational auto-encoders (VAEs) \cite{kingma2013auto} and the Kullback-Leibler divergence \cite{kullback1951information}, we show that the training set with SPDA-generated data is closer to the test set in terms of the overall data distribution.

\subsection{Choosing a superpixel generation method}
Boundary recall and compactness are two key criteria for generation of superpixels. Boundary recall evaluates how well the generated superpixels represent or cover the important object boundaries/contours in an image. Compactness describes how regular and well-organized the superpixels are. Compactness of superpixels tends to constrain superpixels to fit some irregular and subtle object boundaries. In general, one aims to generate superpixels with high boundary recall and high compactness. %With a larger number of superpixels, one may achieve both high boundary recall and good compactness. However too many superpixels can make each superpixel have a too small size and the superpixels be too similar to the original pixels to be actually useful for data augmentation.

For deep learning model training, we aim to generate superpixels with the following properties: (i) good boundary recall, (ii) being compact and pixel-like, and (iii) only pixel values and local image features are used to generate superpixels. Note that many fully convolutional networks work in a bottom-up fashion; superpixels that are generated using global-level information may confuse the training of an FCN model. Hence, we prefer to use superpixel generation method that only utilizes local image information for the pixel grouping process.

SLIC \cite{achanta2012slic} is one of the most widely used methods for generating superpixels. SLIC is fast to compute and can produce good quality superpixels with an option to let the user control the compactness of the generated superpixels. Also, SLIC utilizes only local image information for grouping pixels into superpixels, which is a desired feature by SPDA for training deep learning models. Thus, in our experiments, we use SLIC \cite{achanta2012slic} to generate superpixels for our superpixel-based data augmentation method. The added computational cost for applying SLIC to every training sample is very small comparing to the model training time cost.

\subsection{Generating superpixelized images}\label{gsp}
Suppose the given training set contains $n$ training samples $(x_i, y_i), i = 1,2, \ldots, n$, where $x_i$ is a raw image and $y_i$ is its corresponding annotation map. We apply a superpixel generation method (e.g., SLIC \cite{achanta2012slic}) $F(x_i,s)$ to each image $x_i$ to obtain superpixel cells $c_j^i, j=1,2, \ldots, s$. Each superpixel cell contains a connected set of pixels. Here, $s$ is part of the input to $F$ that specifies the desired number of superpixels that $F$ should produce. We will discuss how to choose the values of $s$ below. Any two different superpixel cells have zero common elements (pixels). The union of the pixels of all the superpixel cells for $x_i$ is all the pixels in the image $x_i$. 

To generate a superpixelized image for $x_i$, for each superpixel cell $c_j^i$, we compute the mean values of all the pixels in $c_j^i$ and update the values of all the pixels in $c_j^i$ using such computed mean values. This step aims to erase low-level pixel variance so that the mid-level and high-level information can be better emphasized by the superpixelized images. We repeat this process for all the superpixel cells of $x_i$, and then form a superpixelized image $SP(x_i,s)$, where $s$ indicates that this superpixelized image is generated using $s$ superpixels. To avoid artificially changing the distribution of annotation (label) maps, the annotation map for $SP(x_i,s)$ is kept as the original $y_i$. Thus, we put $(SP(x_i,s),y_i)$ into our new training data set generated using $(x_i,y_i)$.

The value $s$ specifies the desired number of superpixels to generate. A small number of superpixels would make a superpixelized image too coarse to represent the essential object structures in the original image. A large number of superpixels would make a superpixelized image too similar to the original image. We aim to model a relatively continuous change from each original image sample to its superpixelized images, from fine to coarse, so that the VRM distribution (or neighborhood distribution) around the original image sample can be better captured. As a result, we choose a range $[s_l, s_u]$ of values for $s$, and form a set of superpixelized images $SP(x_i, s)$, $s=s_l, \ldots, s_u$, for each original image $x_i$. 

{\color{black}{For biomedical image datasets, the imaging settings are usually known in practice. In particular, the scales and size of the images, and the range of sizes of objects in the images are often known. Thus, one can set the values of $s_l$ and $s_u$ based on prior knowledge of these image aspects. For different image sets and applications, one can set $s_{l}$ and $s_{u}$ differently. In our experiments, for simplicity and for demonstrating the robustness of SPDA, we choose a common setting of $s_{l}$ and $s_{u}$ for all the 2D segmentation datasets ($s_{l}= 800$ and $s_{u}= 2000$). For the 3D image dataset, due to the increase of image dimensionality, we set $s_{l}= 2000$ and $s_{u}= 4000$.}}

%In order to tackle these difficulties and avoid any bias in choosing the number $s$, we produce multiple versions of superpixelized images for each raw training image (using a range of values for $s$). During training, we randomly choose training data from these versions of superpixelized images for training a segmentation network so that no particular version is favored from our data augmentation point of view. Using $400$ to $800$ superpixels can normally represent the essential object structures in a natural scene image \cite{achanta2012slic,li2015superpixel}. In biomedical images, sometimes objects can be of very small sizes, and there can be many objects of interest in one image. Thus, we choose the values of $s$ to be relatively larger than the cases in natural scene images. We describe this the range of $s$ we used in the experiments section. 

%from $800$ to $2000$ in our experiments. After ``superpixelizing" all the original images in the training set $(x_i,y_i)$, $i=1,2, \ldots,n$, we obtain a new training set as $(x_i^s, y_i)\cup (x_i,y_i)$, $i=1,2, \ldots,n$ and $s=800,900, \ldots, 2000$.

\subsection{Model training using SPDA}\label{trainingUsingSP}

The loss function for training a deep learning based segmentation network using both the original training data and the augmented data is:
\begin{equation}\label{eq6}
\frac{1}{n}\sum_{i=1}^{n}(\mathcal{L}(f(x_i),y_i) +  \lambda \sum_{s=s_{l}}^{s_{u}}\mathcal{L}(f(SP(x_i,s)),y_i))
\end{equation}
\noindent
where $\mathcal{L}$ is a spatial cross-entropy loss, $f$ is the segmentation model under training, $SP$ is for generating a superpixelized image, and $s$ is a parameter for $SP$ that specifies how many superpixels are desired to be generated. We set $\lambda$ as simple as a normalization term $ \frac{1}{s_u-s_l+1}$. We aim to minimize the above function with respect to the parameters of $f$.

A common way of optimizing the objective function above is to use a mini-batch based stochastic gradient descent method. Following the loss function in Eq.~(\ref{eq6}), half of the total samples in the mini-batch is drawn from the original image samples and the other half is from the SPDA-generated samples. We provide the pseudo-code ({\bf Algorithm 1}) for the model training procedure below.\\ %A remaining question is whether in the mini-batch, when $x_i$ is drawn, its corresponding $SP(x_i,s)$ should also be drawn in the same mini-batch.

\begin{algorithm2e}[]
    \KwData{$(x_i, y_i)$ and $(SP(x_i,s),y_i)$, $i=1,2, \ldots, n$ and $s=s_{l}, \ldots, s_{u}$.}
    \KwResult{A trained FCN model.}
    Initialize an FCN model with random weights, mini-batch = $\emptyset$\;
    \While{stopping condition not met}{
        \For{m = 1 to batch-size$/2$}{
            $p=random.randint(1,n)$\;
            add $(x_p,y_p)$ to the mini-batch\;
            $k=random.randint(s_{l},s_{u})$\;
            add $(SP(x_p,k),y_p)$ to the mini-batch\;
        }
        Update FCN using data in the mini-batch using the Adam optimizer\;
        mini-batch = $\emptyset$\;
        }
    \caption{Model training using SPDA-augmented training data}
\end{algorithm2e}

\section{Experiments}\label{sec:3}
Four biomedical image segmentation datasets are used to evaluate our 
%proposed superpixel-based data augmentation (SPDA) 
SPDA method. These datasets are: (1) 3D magnetic resonance (MR) images of myocardium and great vessels (blood pool) in cardiovascular \cite{pace2015interactive}, (2) electron micrographs (EM) of neuronal structures \cite{lee2015recursive},  (3) an in-house 2D electron micrographs (EM) of fungal cells that invade animal (ant) tissues, and (4) 2D H\&E stained histology images of glands \cite{sirinukunwattana2017gland}. {\color{black} {Note that SPDA can be extended to segmentation of 3D images using a straightforward extension of SLIC \cite{achanta2012slic} that generates supervoxels instead of superpixels.}}

On the 2D segmentation datasets, our experiments of SPDA use two common FCN models: U-Net \cite{ronneberger2015u} and DCN \cite{chen2016deep}. In addition to showing the effectiveness of SPDA, on the neuronal structure and fungus datasets, we also compare SPDA with the elastic deformation for data augmentation (EDDA) used in \cite{ronneberger2015u}. On the 3D segmentation dataset, a state-of-the-art DenseVoxNet \cite{yu2017automatic} is utilized for experiments with our SPDA. Experiments on this 3D dataset aim to show the capability of SPDA for 3D image data.

We made a simple extension of the original DCN model \cite{chen2016deep}, which now contains 5 max-pooling layers (deeper than the original DCN). The extension allows DCN to have a larger receptive field for making use of higher-level image information. Random cropping, flipping, and rotation are applied as standard/basic data augmentation operations to all the instances in the experiments. We denote this set of basic data augmentation operations as $DA_{basic}$. For fair comparison, we keep all the training settings (e.g., random seed, learning rate, mini-batch size, etc) the same for all the model training. Adam \cite{kingma2014adam} optimizer is used for model optimization. As in a common practice, the learning rate for model training is set as 0.0005 for the first 30000 iterations, and then decays to 0.00005 for the rest of the training. The mini-batch size is set as 8. The compactness parameter for SLIC \cite{achanta2012slic} is set as its default value 20. The size of the input and output of an FCN model is set as $192\times192$ for 2D images and $64\times64\times64$ for 3D images. {\color{black}{ The training procedure stops its execution when there is no significant change in the training errors.}}

\textbf{3D cardiovascular segmentation}.
The HVSMR dataset \cite{pace2015interactive} was used for segmenting myocardium and great vessels (blood pool) in 3D cardiovascular magnetic resonance (MR) images. The original training dataset contains 10 3D MR images, and the test data consist of another 10 3D MR images. The ground truth of the test data is not available to the public; the evaluations are done by submitting segmentation results to the organizers' server.

%\begin{figure}[h]
%  \centering
%  \includegraphics[width=2.7 in]{heart_sample.pdf}
%  \caption{Left: a cardiovascular 3D MR image; right: the corresponding segmentation ground truth. This is a three-class 3D segmentation problem.}
%  \label{fig_heart}
%\end{figure}

The Dice coefficient, average distance of boundaries (ADB), and symmetric Hausdroff distance are the criteria for evaluating the quality of the segmentation results. A combined score $S$, computed as $S=\sum_{class}(\frac{1}{2}\mbox{\textit{Dice}} - \frac{1}{4}\mbox{\textit{ADB}} -\frac{1}{30}\mbox{\textit{Hausdorff}})$, is used by the organizers, and this score aims to measure the overall quality of the segmentation results.

When applying SPDA to 3D image data, supervoxels (instead of superpixels) are generated. We use a 3D version of SLIC for generating supervoxels. SPDA is tested using the DenseVoxNet~\cite{yu2017automatic}, which is a state-of-the-art FCN for 3D voxel segmentation. In Table \ref{tabHVSMR1}, we show the results from DenseVoxNet + $DA_{basic}$, DenseVoxNet + $DA_{basic}$ + SPDA, and other known models on this dataset.  One can see that SPDA improves the segmentation results significantly, especially on the average distance of boundaries and Hausdroff distance metrics.

\begin{table*}[h]
\centering
\caption{Comparison of segmentation results on the HVSMR dataset.}
\label{tabHVSMR1}
%\tiny

%\footnotesize
\renewcommand\arraystretch{1.3}
% adjust the height of cells of table
\setlength{\tabcolsep}{2pt}
\scriptsize
\begin{tabular}{|c|c|c|c|c|c|c|c|}

    \hline
    %\toprule
    \multirow{2}{*}{ Method } &   \multicolumn{3}{c|}{Myocardium} & \multicolumn{3}{c|}{Blood pool}  &  \multirow{2}{*}{\shortstack{Overall \\ score}} \\  \cline{2-7}
    & Dice   &  ADB    &  Hausdorff    & Dice   &  ADB    &  Hausdorff  &  \\ 
    \hline

    3D U-Net \cite{cciccek20163d} %\footnote{\label{Note1} Obtained from \cite{yu2017automatic}.} \cite{cciccek20163d}   
    &  $0.694$  &  $1.461$  & $10.221$  &   $ 0.926$ & $ 0.940 $  &  $ 8.628$   &  $ -0.419 $ \\ 
    \hline 

    VoxResNet \cite{chen2017voxresnet}  & $ 0.774 $   &  $ 1.026  $  &  $ 6.572  $  &  $ 0.929  $ & $ 0.981 $  & $ 9.966 $     & $ -0.202 $  \\ 
    \hline 

    DenseVoxNet  \cite{yu2017automatic} + $DA_{basic}$ &  $ \bf0.821  $  &  $ 0.964  $  &  $ 7.294  $  & $ 0.931  $  &  $ 0.938  $  &  $ 9.533  $    &  $ -0.161 $ \\ 
    \hline
    DenseVoxNet  \cite{yu2017automatic} + $DA_{basic}$ + SPDA &  $ 0.817  $  &  $ \bf0.723  $  &  $ \bf3.639  $  & $ \bf0.938  $  &  $ \bf0.778  $  &  $ \bf 5.548  $    &  $ {\bf 0.196} $ \\ 
    \hline

     %\midrule
    \hline
\end{tabular}
%}
\end{table*}

%\begin{figure*}[]
%  \centering
%  \includegraphics[width=6.0 in]{dataset_overview.pdf}
%  \caption{From left to right: a neuronal structure image sample, a neuronal structure segmentation ground truth sample, a fungus image sample, a fungus segmentation ground truth sample (colors indicate object classes), a gland image sample, a gland segmentation ground truth sample (colors indicate gland object instances).}
%\label{fig:dataset_overview}
%\end{figure*}
\textbf{Fungal segmentation}. We further evaluate SPDA using an in-house EM fungus dataset that contains 6 large 2D EM images ($4000 \times 4000$ each) for segmentation.  Since the input window size of a fully convolutional network is set as $192 \times 192$, there are virtually hundreds and thousands unique image samples for model training and testing. This dataset contains three classes of objects of interest: fungal cells, muscles, and nervous tissue. We use 1 large microscopy image for training, and 5 large microscopy images for testing.  This experiment aims to evaluate the effectiveness of SPDA in a difficult situation in which the training set is smaller than the test set (not uncommon in biomedical image segmentation). {\color{black}{In Table \ref{fungus_results}, Student's t-test suggests that all our improvements are significant. The p-values for MeanIU of U-Net vs U-Net + SPDA, U-Net + EDDA vs U-Net + SPDA, DCN vs DCN + SPDA, and DCN + EDDA vs DCN + SPDA are all $<$ 0.0001.}}

%\begin{figure}[h]
%  \centering
%  \includegraphics[width=2.1 in]{fungus_dataset_sample.pdf}
%  \caption{Left: a raw EM image of fungal cells and tissues; right: the corresponding segmentation ground truth (colors mark object classes). It is a four-class segmentation problem.}
%\label{fig:dataset_overview}\label{fig_fungus}
%\end{figure}

\begin{table*}[h]
\centering
	\caption{Comparison results on the fungus segmentation dataset: The Intersection-over-Union (IoU) scores for each object class and the MeanIU scores across all the classes of objects. U-Net \cite{ronneberger2015u}, DCN \cite{chen2016deep}, and EDDA: elastic deformation data augmentation (used in \cite{ronneberger2015u}) are considered.
	}
	\label{fungus_results}
\scriptsize
\begin{tabular}{|c|c|c|c|c|}
  \hline
  {Method}&{Fungus}&{Muscle}&{Nervous tissue}& {MeanIU}\\
    \hline
U-Net + $DA_{basic}$ & {0.849 $\pm$ 0.008 } & {0.976 $\pm$ 0.003} & {0.506 $\pm$ 0.029} & 0.777 $\pm$ 0.008\\\hline
U-Net + $DA_{basic}$ + EDDA & {0.881 $\pm$ 0.007 } & {0.975 $\pm$ 0.004} & {0.549 $\pm$ 0.035} & 0.8019 $\pm$ 0.014\\\hline
U-Net + $DA_{basic}$ + SPDA & {0.927 $\pm$ 0.001} & {0.973 $\pm$ 0.002} & {0.667 $\pm$ 0.020}& \textbf{0.856 $\pm$ 0.007}\\
  \hline
   \hline
    DCN + $DA_{basic}$ & $0.783\pm0.064$ & $0.970 \pm 0.009$ & $0.349 \pm 0.092$& $0.701 \pm 0.055$\\\hline
  
   DCN + $DA_{basic}$ + EDDA & $0.863\pm0.042$ & $0.970\pm0.008$ & $0.453\pm0.183$  & $0.762\pm0.078$\\\hline
    DCN + $DA_{basic}$ + SPDA & {0.907 $\pm$ 0.011} & {0.973 $\pm$ 0.005} & {0.630 $\pm$ 0.026} & \textbf{0.837 $\pm$ 0.012}\\  \hline

\end{tabular}
\end{table*}

\textbf{Neuronal structure segmentation}. We experiment with SPDA using the EM mouse brain neuronal images \cite{lee2015recursive}. This dataset contains 4 stacks of EM images (1st: $255\times255\times168$, 2nd: $512\times512\times170$, 3rd: $512\times512\times169$, and 4th: $256\times256\times121$). Following the practice in \cite{lee2015recursive,shen2017multi}, we use the 2nd, 3rd, and 4th stacks for model training and the 1st stack for testing. {\color{black}{Since the image stacks in this dataset are highly anisotropic (i.e., the voxel spacing along the $z$-axis is much larger than those along the $x$- and $y$-axes), directly applying 3D models with 3D convolutions is not very suitable for highly anisotropic 3D images. Hence, for simplicity, our experiments on this dataset are based on superpixels in the 2D slices of the 3D images and using 2D FCN models, instead of supervoxels and 3D models.}} We run all experiments 5 times with different random seeds. The average performance across all the runs and their standard deviations are reported in Table \ref{neuro_results}. {\color{black}{Student's t-test suggests that all our improvements are significant. The p-values for $V_{Fscore}^{Rand}$ are: $<$ 0.0001 for U-Net vs U-Net + SPDA, 0.0059 for U-Net + EDDA vs U-Net + SPDA, $<$ 0.0001 for DCN vs DCN + SPDA, and 0.0042 for DCN + EDDA vs DCN + SPDA.}}

%\begin{figure}[h]
%  \centering
%  \includegraphics[width=2.1 in]{neuro_dataset_sample.pdf}
%  \caption{Left: a raw image of neuronal structures; right: the corresponding segmentation ground truth. This is a two-class segmentation problem.}
%\label{fig_neuronal}
%\end{figure}
 % \begin{figure}[]

   %     {\includegraphics[width=8 cm]{neuroFigure_thin_cropped.pdf}}{\caption{On the neuronal structure segmentation dataset: $V^{Rand}$ scores for $M^2$ FCN \cite{shen2017multi}, SN, SN with elastic deformation data augmentation (EDDA), and SN with our proposed superpixel-based data augmentation (SPDA).}\label{fig:neuroFigure}}

%\end{figure}
        
        %  \begin{figure}[]
        %{\includegraphics[width=4.6 cm]{fungus_exp_samples_new}}{\centering \caption{Fungus related segmentation result samples: (a) a raw image region, (b) by SN, (c) by SN with EDDA, (d) by SN with SPDA, (e) the ground truth image region (red: fungal cells, green: muscles, blue: nervous tissue).}\label{fig:fungus_exp_samples}}
        %\end{figure}
%
%\begin{figure}[t]
%  \centering
%  \hspace{-1 cm}
%  \includegraphics[width=3.4 in]{neuroFigure_cropped.pdf}
%  \caption{On the neuronal structure segmentation dataset: $V^{Rand}$ scores for $M^2$ FCN \cite{shen2017multi}, SN, % (an encoder-decoder based FCN model from \cite{zhang2017deep}), 
%  SN with elastic deformation data augmentation (EDDA), and SN with our proposed superpixel-based data augmentation (SPDA).}
%\label{fig:neuroFigure}
%\end{figure}

\begin{table*}[h]
\centering
\caption{Comparison results on the neuronal structure segmentation dataset: $V^{Rand}$ scores for evaluating the segmentation quality. $DA_{basic}$: basic data augmentation operations (random cropping, flipping, and rotation); EDDA: elastic deformation data augmentation in \cite{ronneberger2015u}.}
\label{neuro_results}
\scriptsize
\begin{tabular}{|c|c|c|c|}
  \hline
  {Method}&{$V_{merge}^{Rand}$}&{$V_{split}^{Rand}$}&{$V_{Fscore}^{Rand}$} \\
    \hline
      $M^2$FCN \cite{shen2017multi} & $0.9917$ & $0.9815$ & $0.9866$\\\hline
      \hline
     U-Net \cite{ronneberger2015u} + $DA_{basic}$ & $0.9954\pm 0.0003 $ & $0.9879\pm 0.0001$ & $0.9917\pm 0.0001$\\\hline
      U-Net + $DA_{basic}$ + EDDA & $0.9957\pm 0.0005$ & $ 0.9931\pm 0.0003$ & $0.9944\pm 0.003$ \\\hline
     U-Net + $DA_{basic}$ + SPDA & \textbf{0.9965 $\pm$ 0.0003} & \textbf{0.9935 $\pm$ 0.0004} & \textbf{0.9950 $\pm$ 0.0002}\\  
      \hline
      \hline
  DCN \cite{chen2016deep} + $DA_{basic}$ & $0.9950\pm0.0003$ & $0.9916\pm0.0001$ & $0.9933\pm0.0001$\\\hline
       DCN + $DA_{basic}$ + EDDA & $0.9980\pm 0.0006$ & $0.9917\pm0.0001$ & $0.9949\pm 0.0002$ \\\hline
    DCN + $DA_{basic}$ + SPDA & \textbf{0.9987 $\pm$ 0.0010} & \textbf{0.9921 $\pm$ 0.0006} & \textbf{0.9954 $\pm$ 0.0002}\\
  
  \hline
  
\end{tabular}
\end{table*}

\textbf{Gland segmentation}. This H\&E stained microscopy image dataset \cite{sirinukunwattana2017gland} contains 85 training images (37 benign (BN), 48 malignant (MT)) and 60 testing images (33 BN, 27 MT) in part A, and 20 testing images (4 BN, 16 MT) in part B. Table \ref{gland_results_1} shows the gland segmentation results that demonstrate the effect of SPDA and comparison with the state-of-the-art models on this dataset. In particular, using SPDA, DCN can be trained to perform considerably better than the state-of-the-art model \cite{graham2018mild}.

%\begin{figure}[h]
%  \centering
%  \includegraphics[width=2.4 in]{gland_dataset_sample.pdf}
%  \caption{Left: a raw image of glands; right: the %corresponding segmentation ground truth. This is a %two-class segmentation problem.}
%\label{fig_gland}
%\end{figure}

\begin{table*}[!]
\small
\centering
	\caption{Comparison results on the gland segmentation dataset: The $F_1$ score and ObjectDice evaluate how well glands are %detected and 
	segmented at the instance level, and Object Hausdorff distance evaluates the shape similarity between the segmented objects and ground truth objects.}
	\label{gland_results_1}
	\scriptsize
\begin{tabular}{|c|c|c|c|c|c|c|}

  \hline
  \multirow{2}{*}{Method}&\multicolumn{2}{c|}{$F_1$ Score}&\multicolumn{2}{c|}{ObjectDice}&\multicolumn{2}{c|}{ObjectHausdorff}\\\cline{2-7}
  &\multicolumn{1}{c|}{part A}&\multicolumn{1}{c|}{part B}&\multicolumn{1}{c|}{part A}&\multicolumn{1}{c|}{part B}&\multicolumn{1}{c|}{part A}&\multicolumn{1}{c|}{part B}\\
    \hline
    CUMedVision \cite{chen2016dcan}& 0.912 & 0.716 & 0.897  &0.718  &45.418  &160.347 \\
  \hline
    Multichannel1 \cite{xu2016glandmiccai}& 0.858  & 0.771  & 0.888  &0.815& 54.202&129.930 \\
  \hline

  Multichannel2 \cite{xu2016gland}& 0.893  & 0.843  & 0.908  &0.833&{44.129}&116.821 \\
   \hline
   MILD-Net \cite{graham2018mild} &0.914 & {0.844} & \textbf{0.913} &0.836&\textbf{41.54}&105.89  \\
  \hline
\hline
  U-Net \cite{ronneberger2015u} + $DA_{basic}$ & 0.89202 & 0.8087 & 0.88193  &0.83441 &{51.19}  &{108.25} \\\hline
  U-Net + $DA_{basic}$ + SPDA &{0.9007} & {0.83843} & {0.88429} &{0.8415}&{49.95}&{107.69}  \\
  \hline
  \hline

DCN \cite{chen2016deep} + $DA_{basic}$ & 0.9071 & 0.825 & 0.898  &0.826 &48.740  &126.479 \\\hline
  DCN + $DA_{basic}$ + SPDA &\textbf{0.918} & \textbf{0.860} & \textbf{0.913} &\textbf{0.858}&{42.620}&\textbf{95.83}  \\
  \hline

\end{tabular}
\end{table*}

\section{Conclusions}
In this paper, we presented a new data augmentation method using superpixels (or supervoxels), SPDA, for training fully convolutional networks for biomedical image segmentation. Our proposed SPDA method is well motivated, easy to use, compatible with known data augmentation techniques, and can effectively improve the performance of deep learning models for biomedical image segmentation tasks.

\bibliography{SPDA}

\appendix
\section{Empirical Studies of SPDA-generated Data} \label{EmpiricalStudy}
In this appendix, two empirical studies are conducted to show: (1) the SPDA-generated data are ``near" their original image data in the data space, and (2) the data distribution of the SPDA-augmented training set is ``closer'' to the distribution of the test data (or true data).

\subsection{SPDA-generated data near their original samples}\label{EmpiricalStudy-1}
 We seek to examine how the SPDA-generated data are spatially close to their corresponding original images. Since $SP(x_i,s)$ and $x_i$ are both in a high dimensional space, comparing them is not a trivial task. One may use a distance metric for measuring the distance between $SP(x_i,s)$ and $x_i$. However, with different metrics, the meaning of ``being different" or ``being similar" can be drastically different.
 
\begin{figure*}[!]
	\centering
	\includegraphics[width=15cm]{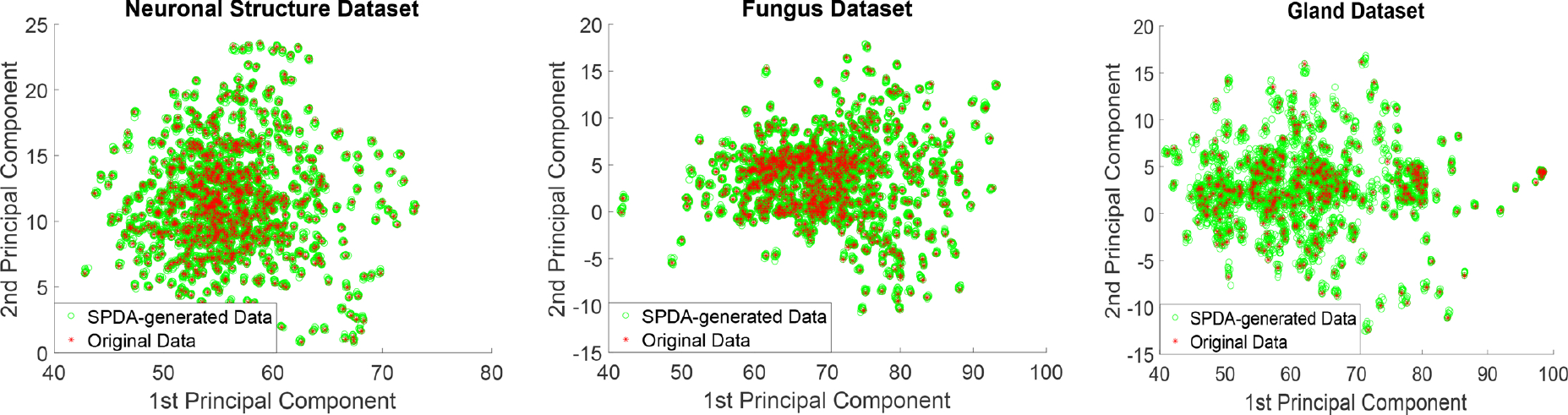}
	\caption[]{After dimensionality reduction using PCA, each original image sample (red) is surrounded by (or closely adjacent to) its corresponding superpixelized images (green). Zoom-in view would show more details.}
	\label{fig_PCA_ALL}
\end{figure*}

\begin{figure*}[!]
	\centering
	\includegraphics[width=15 cm]{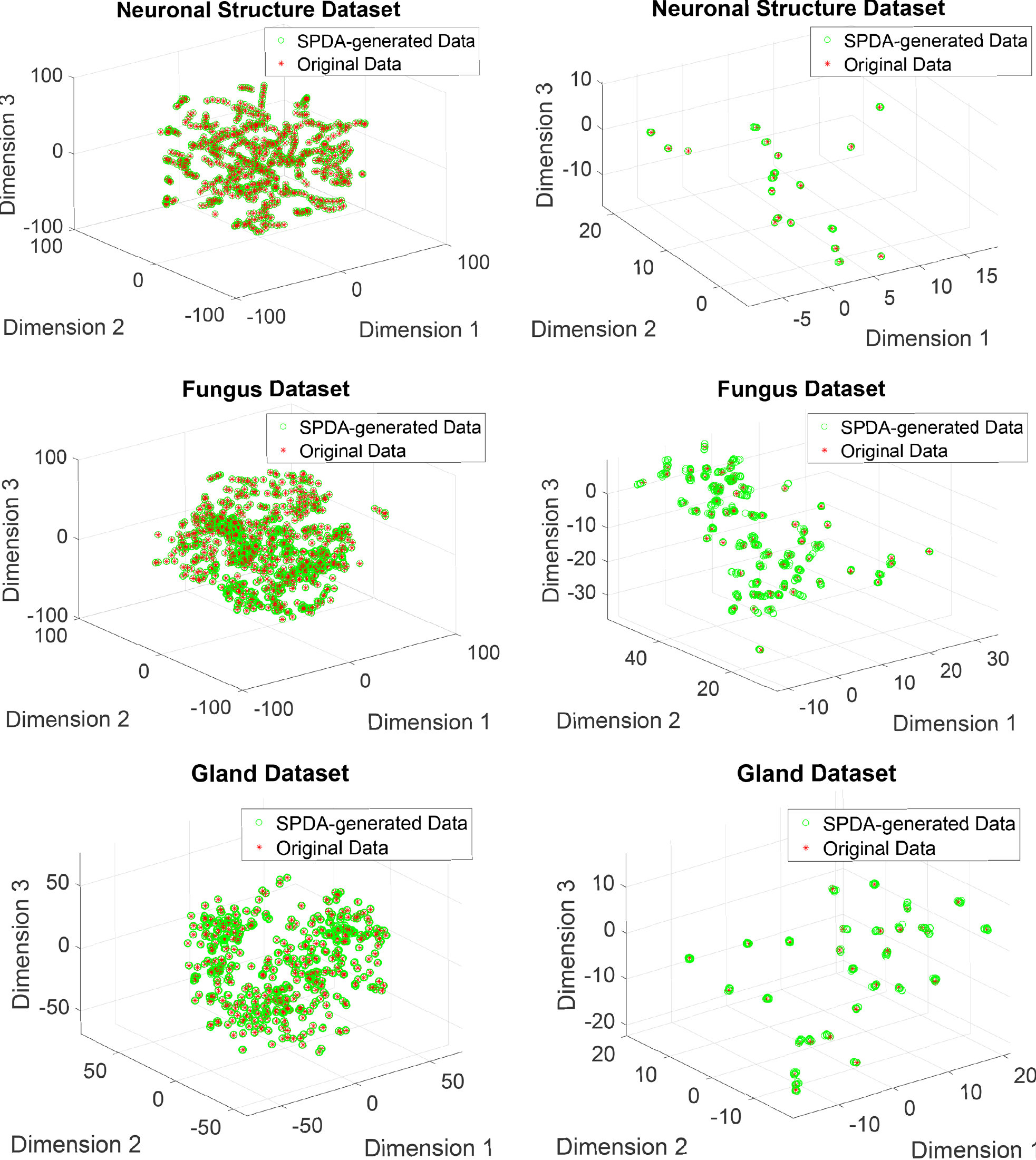}
	\caption[]{Visualization of some SPDA-generated image samples (green) and the original training samples (red) using t-SNE \cite{maaten2008visualizing}. Left: Overview of the samples; right: zoom-in views. SPDA-generated samples are in a close neighborhood of their corresponding original samples.}
	\label{tSNE}
\end{figure*}

 To avoid too much complication in manifold learning or metric learning, we use two common dimensionality reduction methods, standard PCA and t-SNE \cite{maaten2008visualizing}, to help visualize the original image samples and SPDA-generated image samples. Fig.~\ref{fig_PCA_ALL} shows visualization results of such samples (both the original and SPDA-generated samples) on the neuronal structure dataset, fungus dataset, and gland dataset (after applying PCA). One may observe that the SPDA-generated data are near/surrounding the original image data, forming a close neighborhood of the original images.  In Fig.~\ref{tSNE}, we provide visualization views of some SPDA-generated image samples using t-SNE \cite{maaten2008visualizing}.
 
\subsection{Data distribution comparison}\label{EmpiricalStudy-2}
Here we are interested in a basic question: Whether adding SPDA-generated data $X_{spda}$ to the original training set $X_{ori}$ makes the new training set $X_{augmented}$ ``closer'' to the test data $X_{test}$ in the image representation space.

We utilize variational auto-encoders (VAEs) \cite{kingma2013auto} to encode the training images $X=\{x_1, \ldots,x_n\}$ into much lower dimensional representation $Z=\{z_1, \ldots,z_n\}$. On each dimension of the space thus resulted, the data are expected to follow a Gaussian distribution with zero mean and unit variance. This is a standard objective of VAE.

\begin{figure}[!]
	\centering
	\includegraphics[width=7cm]{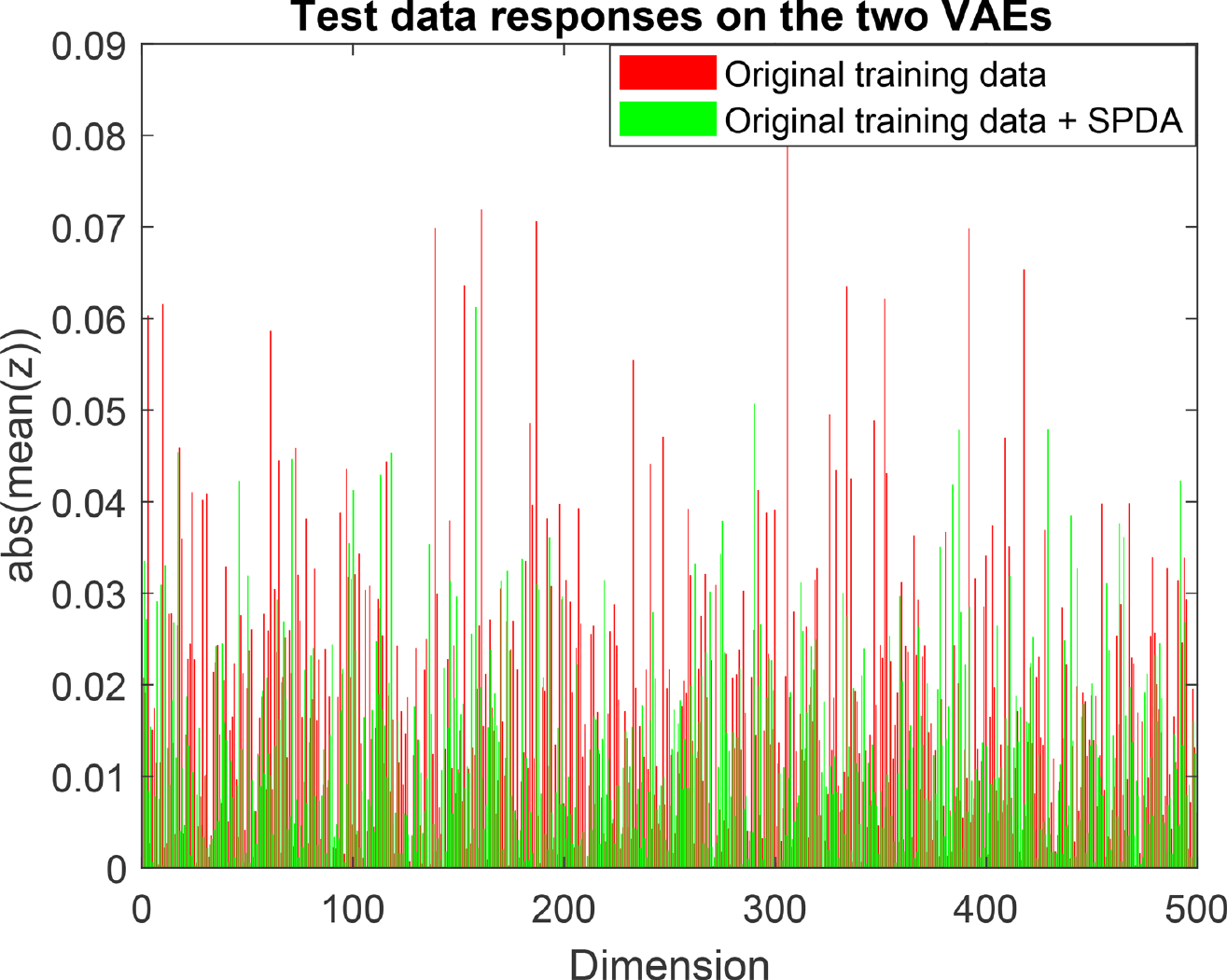}
	\caption[]{The responses of the test data on the two VAEs trained using the original training data and the SPDA-augmented training data (on the neuronal structure dataset).}
	\label{figvae}
\end{figure}

To show the effect of SPDA, we train two VAEs: VAE-A is trained using only the original training images $X_{ori}$, and VAE-B is trained using the SPDA-augmented training set $X_{augmented} =  X_{ori} \cup X_{spda}$. These two VAEs are all trained using the same settings; the only difference is their training data. After training, VAE-A is applied to its training data $X_{ori}$ and the test data $X_{test}$, to obtain $Z_{ori}^{A}$ and $Z_{test}^{A}$. Similarly, VAE-B is applied to its training data $X_{augmented}$ and the test data $X_{test}$, and $Z_{augmented}^{B}$ and $Z_{test}^{B}$ are obtained. We then compare 
\begin{equation}
D_{KL}(P(z_{test}^{A})||P(z_{ori}^{A}))
\end{equation}
\noindent
with 
\begin{equation}
D_{KL}(P(z_{test}^{B})||P(z_{augmented}^{B}))
\end{equation}
\noindent
and compare 

\begin{equation}
D_{KL}(P(z_{ori}^{A})||P(z_{test}^{A}))
\end{equation}
\noindent
with 
\begin{equation}
D_{KL}(P(z_{augmented}^{B})||P(z_{test}^{B}))
\end{equation}
\noindent
where $D_{KL}$ is the Kullback-Leibler divergence \cite{kullback1951information} and $P(z)$ is the probability distribution of $z$. The above procedure is applied to the neuronal structure dataset. The results are: {%\scriptscriptstyle
${D_{KL}(P(z_{test}^{B})||P(z_{augmented}^{B}))}=5.5517 $$<$${D_{KL}(P(z_{test}^{A})||P(z_{ori}^{A}))}=5.8779$, and ${D_{KL}(P(z_{augmented}^{B})||P(z_{test}^{B}))}=5.0586 < {D_{KL}(P(z_{ori}^{A})||P(z_{test}^{A}))}=6.1491$}. It is clear that SPDA can potentially make the training data distribution closer to the test data/true data distribution in the image representation space. We believe this is a main reason why learning models trained using SPDA-augmented training data can generalize better on test data. To show this observation visually, the absolute values of the averages of $Z_{test}^{A}$ and $Z_{test}^{B}$ are shown in Fig~\ref{figvae}. One can see that the values of $Z_{test}^{B}$ are generally closer to 0 than $Z_{test}^{A}$, which means that the distribution of $Z_{test}^{B}$ is closer to the zero mean Gaussian distribution.

\end{document}